\documentclass{article}

 \PassOptionsToPackage{numbers, compress}{natbib}
\usepackage[final]{nips_2017}

\usepackage[utf8]{inputenc} 
\usepackage[T1]{fontenc}    
\usepackage{hyperref}       
\usepackage{url}            
\usepackage{booktabs}       
\usepackage{amsfonts}       
\usepackage{amsmath}
\usepackage{nicefrac}       
\usepackage{microtype}      
\usepackage{subfiles}
\usepackage{xcolor}
\usepackage{multirow}
\usepackage{enumerate}
\usepackage{subfiles}
\usepackage{multirow}
\usepackage{graphicx}
\usepackage{subfiles}

\title{Intermediate direct preference optimization}

\author{
	Atsushi Kojima\\
	Money Forward, Inc.\\
	\texttt{kojima.atsushi@moneyforward.co.jp}
}

\begin{document}

\maketitle

\begin{abstract}
We propose the intermediate direct preference optimization (DPO) method to calculate the DPO loss at selected intermediate layers as an auxiliary loss for fine-tuning large language models (LLMs). The conventional DPO method fine-tunes a supervised fine-tuning (SFT) model by calculating the DPO loss using logits from the final layer. In our intermediate DPO approach, DPO losses are calculated using the logits from K-selected intermediate layers and averaged to obtain the intermediate DPO loss. For training the intermediate DPO model, the final loss is obtained by calculating the weighted sum of the DPO and intermediate DPO losses. During inference, the intermediate DPO model decodes using the final layer logits similarly to the conventional DPO model. In experiments using the ultrafeedback dataset, the performance of the intermediate DPO model was evaluated using GPT-4. As a result, the intermediate DPO model trained using the intermediate DPO loss calculated at the 22nd layer of a 32-layer SFT model achieved win rates of 52.5\% and 67.5\% against the conventional DPO and SFT models, respectively, demonstrating the effectiveness of the proposed method. Furthermore, we report the relationships among the position of the selected intermediate layers, the number of layers, and performance.
\end{abstract}

\section{Introduction}
\label{sec:intro}
Preference optimization methods \cite{dpo, ppo1, ppo2, ppo3, ppo4}, which align large language models (LLMs) with human preference, have been recently explored. Among such methods, direct preference optimization (DPO) is the most promising and efficient method, because it can fine-tune a model without training a reward model or complex reinforcement algorithms such as proximal policy optimization \cite{ppo4}.

A DPO model can be obtained by fine-tuning a supervised fine-tuning (SFT) model based on DPO loss. The DPO method encourages the SFT model to learn that the desirable completion is preferred over the rejected completion for a given prompt. DPO loss is calculated using logits obtained from the final layer of the SFT model. 

In this paper, we propose the intermediate DPO method to calculate the DPO loss at selected intermediate layers as an auxiliary loss. In our intermediate DPO approach, DPO losses are calculated using the logits from K-selected intermediate layers and averaged to obtain the intermediate DPO loss. For training the intermediate DPO model, the final loss is obtained by calculating the weighted sum of the DPO and intermediate DPO losses. The parameters of the linear layers used to convert the hidden vectors to the logits are shared at each intermediate layer.
In experiments using the ultrafeedback dataset\footnote{HuggingFaceH4/ultrafeedback\_binarized}, we evaluated the performance of the intermediate DPO model using GPT-4. The results showed that the intermediate DPO model trained using the intermediate DPO loss calculated at the 22nd layer of a 32-layer SFT model achieved win rates of 52.5\% and 67.5\% against the conventional DPO and SFT models, respectively, demonstrating the effectiveness of the proposed method.

\section{DPO}
\label{sec:itr}
The DPO model is obtained by fine-tuning an SFT model. The DPO loss formula is as follows Equation~(\ref{eq:dpo}):
\begin{equation}
	\label{eq:dpo}
	\lambda_{\mathrm{DPO}} = -\log \sigma (\beta \log \frac{\pi_\theta(y_w | x)}{\pi_\text{ref}(y_w | x)} - \beta \log \frac{\pi_\theta(y_l | x)}{\pi_\text{ref}(y_l | x)}).
\end{equation}
Here, $x$ is the prompt, and $y_l$ and $y_w$ represent the undesirable and desirable completions, respectively. $\pi_\theta$ and $\pi_\text{ref}$ represent the optimal and reference policies (SFT), respectively. $\sigma$ is the sigmoid function, and $\beta$ represents a hyperparameter.

\section{Intermediate DPO}
\label{sec:str}
\begin{figure}[htb]
	\centering
	\includegraphics[scale=0.6]{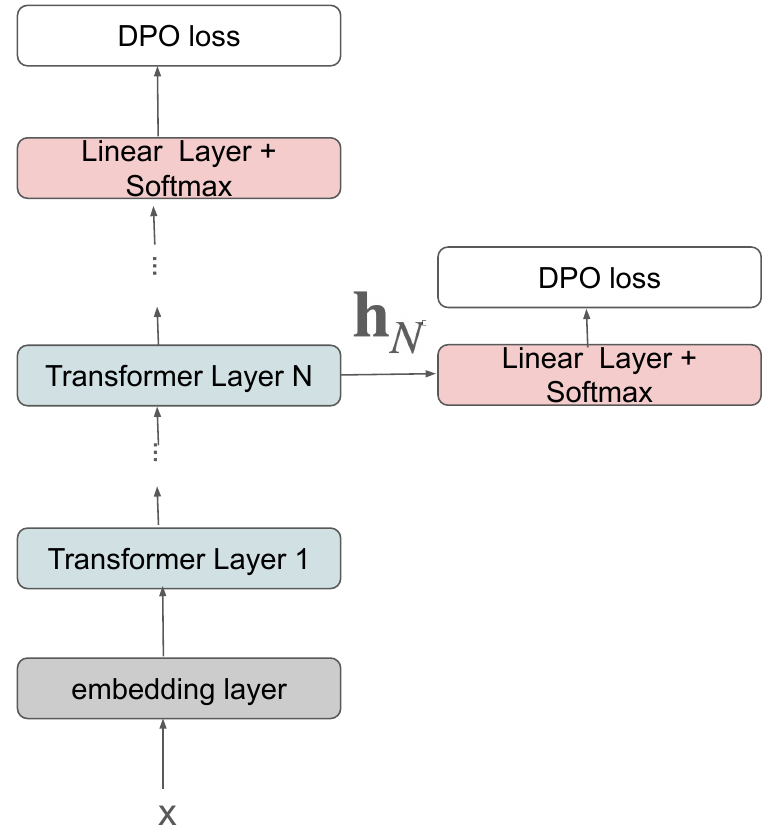} 
	\caption{Overview of intermediate DPO}
	\label{fig:idpo}
\end{figure}

In this paper, we propose calculating the DPO loss at intermediate layers as an auxiliary loss to enhance the performance of the DPO model. The conceptual overview of the proposed method is shown in Figure~\ref{fig:idpo}. In the proposed method, logits are calculated from the hidden vectors of intermediate layers and then the DPO loss is computed. The parameters of the linear layers used to convert the hidden vectors to the logits at each intermediate layer are shared.

The formula for the intermediate DPO loss is shown in Equation~(\ref{eq:interdpo}), where $K$ denotes the indices of the selected intermediate layers. The intermediate DPO loss is obtained by averaging the DPO losses computed at the intermediate layers. For example, if $K=5, 8$, the DPO loss is calculated by using logits from the 5th and 8th layers and then averaging these values.
\begin{equation}
	\label{eq:interdpo}
	\lambda_{\mathrm{DPO}_\mathrm{intermediate}} = \frac{1} {|K|}\sum_{k \in K} \lambda_{\mathrm{DPO}_k}.
\end{equation}

To train the model using the intermediate DPO loss as an auxiliary loss function, we compute the weighted sum of DPO loss. The final loss for training the model is shown in Equation~(\ref{eq:loss}), where $\gamma$ is a hyperparameter.
\begin{equation}
	\label{eq:loss}
	\lambda = \gamma \lambda_{\mathrm{DPO}} + (1-\gamma) \lambda_{\mathrm{DPO}_\mathrm{intermediate}}.
\end{equation}

During inference, the intermediate DPO model calculates posterior probabilities from the logits of the output layer and decodes similarly to the conventional DPO model.

\section{Experiments}
\label{sec:evaluation}
\subsection{Experimental Conditions}
\label{sec:dataset}
All experiments were conducted using the ultrafeedback dataset. For the training of the intermediate DPO model, the entire training dataset was used, and for evaluation, 40 randomly selected prompts from the test dataset were utilized. The evaluation was conducted using GPT-4.

The intermediate DPO model was trained by fine-tuning a 7B SFT model (32-layer model)\footnote{mistralai/Mistral-7B-Instruct-v0.1} by low-rank adaptation (LoRA) \cite{lora}. The training conditions for the intermediate DPO model are shown in Table~\ref{table:params}. The hyperparameter in the proposed method was set at $\gamma=0.9$.
\begin{table}[htb]
	\caption{Training conditions for intermediate DPO model}
	\label{table:params}
	\centering
	\begin{tabular}{|c|l|} \hline
		Parameter & Value \\ \hline
		Adapter position & k\_proj, q\_proj, v\_proj, \\
		& o\_proj, gate\_proj, \\
		& down\_proj, up\_proj \\ \hline
		Lora attention dimension& 16 \\ \hline
		$\alpha$ for LoRA scaling & 16 \\ \hline
		Dropout rate for LoRA & 0.05 \\ \hline 
		Learning rate & 5e-4 \\ \hline
		$\beta$ & 0.1 \\ \hline
		$\gamma$ & 0.9 \\ \hline 
		Batch size & 32 \\ \hline
	\end{tabular}
\end{table}

Inference was performed by temperature sampling. We set to a temperature of 0.3 and a repetition penalty of 1.1.

\subsection{Results}
\label{sec:result}
\begin{figure}[htb]
	\centering
	\includegraphics[scale=0.45]{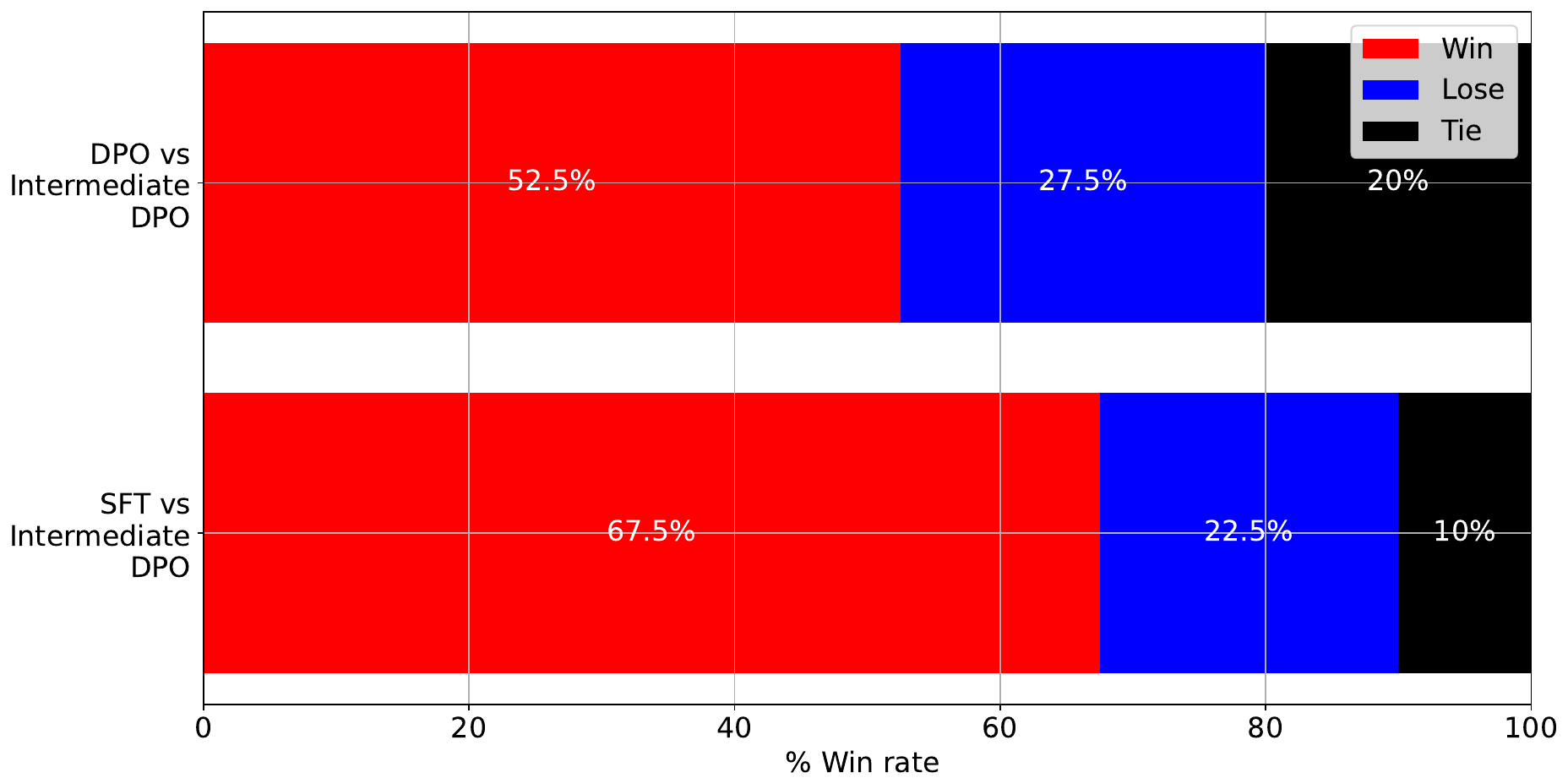}
	\caption{Win rates of the intermediate DPO model against the SFT model and DPO model ($K=11, 22$)}
	\label{fig:main_results}
\end{figure}

Figure~\ref{fig:main_results} shows the win rates of the intermediate DPO model against the SFT and DPO models. The layers at which the intermediate DPO loss was calculated were set to $K=11, 22$. The win rates of the intermediate DPO model against the SFT and DPO models were 67.5\% and 52.5\%, respectively, indicating that the intermediate DPO model outperforms the conventional DPO model.

Table~\ref{table:layer_affect} shows the impact of the layer depth at which the intermediate DPO loss is calculated on accuracy. The numbers in the table indicate the win rates of the intermediate DPO model against the SFT model. The performance of the model in calculating the DPO loss at layers closer to the output (id5) was observed to be higher than those in calculating the DPO loss in the middle (id9) or closer to the input layers (id4), suggesting that calculating the intermediate DPO loss closer to the output layer improves performance.
\begin{table}[htb]
	\caption{Impact of layer depth at which the intermediate DPO loss is calculated on accuracy (32-layer model)}
	\label{table:layer_affect}
	\centering
	\begin{tabular}{|l|l|l|l|l|} \hline
		Exp ID & $K$ & Win rate & Lose rate & Tie \\ \hline \hline 
		id4 & $11$ & 52.5 & 27.5 & 20 \\ \hline 
		id9 & $16$ & 50 & 27.5 & 22.5 \\ \hline 
		id5 & $22$ & 60 & 15 & 25 \\ \hline 
	\end{tabular}
\end{table}

Table~\ref{table:selection_multi} shows the impact of the selection method on the accuracy when calculating DPO loss from multiple intermediate layers. The numbers indicate the win rates of the intermediate DPO model against the SFT model. Comparing the results of selecting from one layer (id4, id5, and id9) with those from two layers (id3 and id10), shows that increasing the number of layers from one to two improves the win rate, indicating that selecting multiple layers to calculate intermediate DPO loss is beneficial.

\begin{table}[htb]
\caption{Impact of selection methods on accuracy}
\label{table:selection_multi}
\centering
\begin{tabular}{|l|l|l|l|l|} \hline
	Exp ID & $K$ & Win rate & Lose rate & Tie \\ \hline \hline 
	id4 & $11$ & 52.5 & 27.5 & 20 \\ \hline 
	id5 & $22$ & 60 & 15 & 25 \\ \hline 
	id9 & $16$ & 50 & 27.5 & 22.5 \\ \hline                     
	id3 & $11, 22$ & 67.5 & 22.5 & 10 \\ \hline         
	id10 & $16, 22$ & 67.5 & 22.5 & 10 \\ \hline                 
\end{tabular}
\end{table}   

Furthermore, since calculating intermediate DPO loss on all layers is computationally expensive, the effect of a layer selection method on the accuracy was investigated. Table~\ref{table:selection_methods} shows a comparison of two methods of selecting intermediate layers: selecting dispersed layers across the network (id3) and choosing consecutive layers closer to the output (id7). The numbers in the table indicate the win rates of the intermediate DPO model against the SFT model. The numbers indicate the win rates of the intermediate DPO model against the SFT model. Selecting from spread out layers (id3) performed better than selecting continuously from layers closer to the output (id7), indicating that a spread out selection method is effective in the proposed intermediate DPO approach.
	\begin{table}[htb]
		\caption{Impact of layer selection methods on accuracy}
		\label{table:selection_methods}
		\centering
		\begin{tabular}{|l|l|l|l|l|} \hline
			Exp ID & K & Win rate & Lose rate & Tie \\ \hline \hline 
			id3 & $11, 22$ & 67.5 & 22.5 & 10 \\ \hline         
			id7 & $30, 31$ & 65 & 22.5 & 12.5 \\ \hline                         
		\end{tabular}
	\end{table}     
	
	\subsection{Discussion}
	\label{sec:discuss}
	To verify the effect of the proposed method, we analyzed the likelihood of DPO across all layers of the intermediate DPO model given by $\log \sigma (\beta \log \frac{\pi_\theta(y_w | x)}{\pi_\text{ref}(y_w | x)} - \beta \log \frac{\pi_\theta(y_l | x)}{\pi_\text{ref}(y_l | x)})$. The analysis was conducted by randomly selecting 30 prompts from the ultrafeedback test dataset and calculating their average.
	
\begin{figure}[htb]
		\centering
		\includegraphics[scale=0.45]{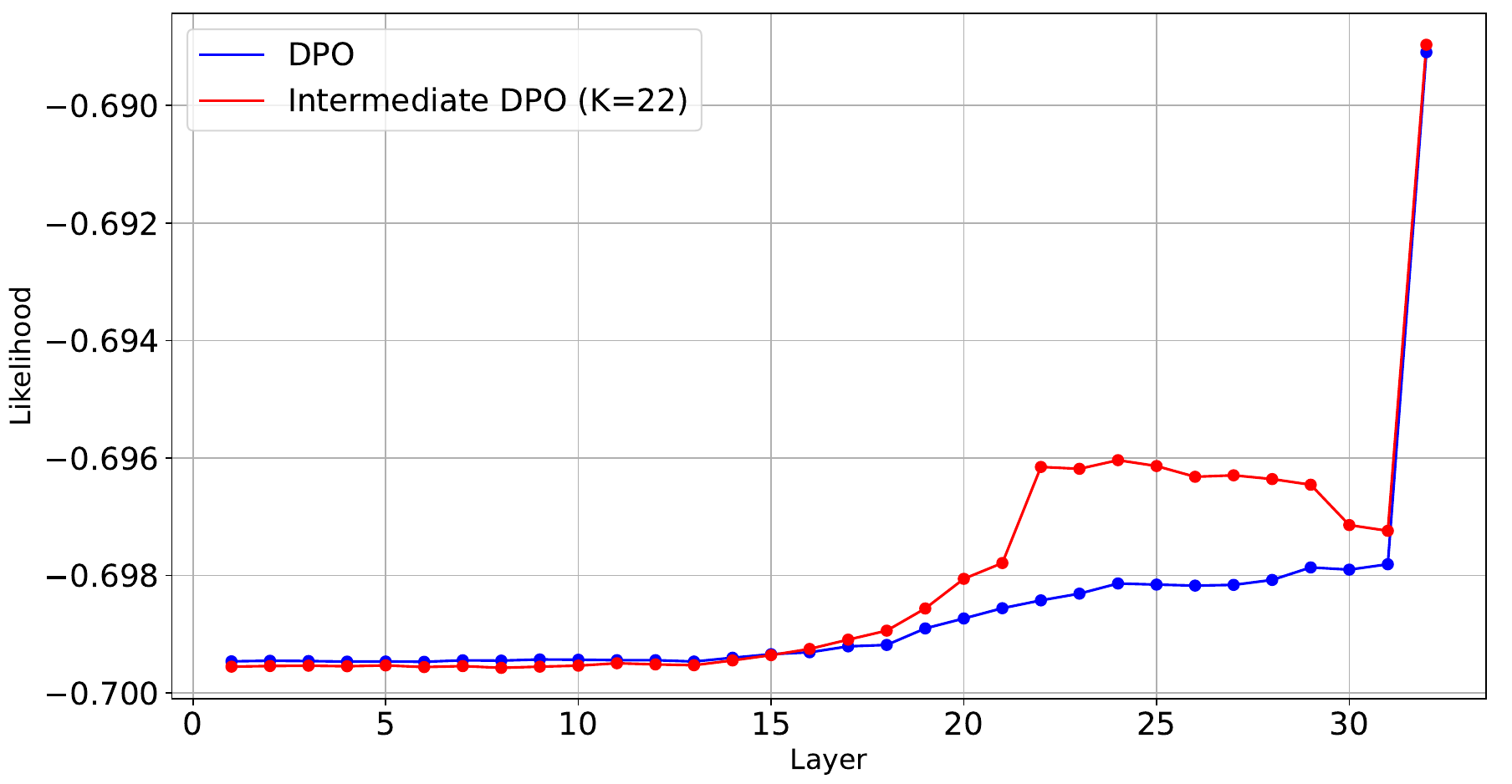} 
		\caption{Comparison of likelihood from intermediate layers}
		\label{fig:analysis}
	\end{figure}

	Figure~\ref{fig:analysis} shows a comparison of likelihoods between the DPO model and the intermediate DPO model (id5) trained with settings $K$ to 22.
	
	In this figure, the horizontal axis represents the intermediate layer indices and the vertical axis represents the DPO likelihood. Since in the proposed method the intermediate DPO loss is calculated at the 22nd layer, a peak can be observed at this layer. Furthermore, owing to the effect of the proposed method, the likelihood in the output layer of the intermediate DPO model is higher than that of the DPO model, which is considered to contribute to the improved performance.
	
	\section{Conclusion}
	\label{sec:conclusion}
	In this paper, we proposed calculating the DPO loss at intermediate layers as an auxiliary loss when fine-tuning LLMs based on DPO, referred to as intermediate DPO. The intermediate DPO loss is obtained by calculating and averaging the DPO losses using logits from the selected K-intermediate layers. When training the model, the final loss is obtained by computing the weighted sum of the DPO and intermediate DPO losses. In experiments using the ultrafeedback dataset, the performance of the intermediate DPO model was evaluated using GPT-4. The results showed that the proposed method achieved win rates of 52.5\% against the DPO model and 67.5\% against the SFT model, demonstrating the effectiveness of the proposed method. In our future work, we will investigate the effectiveness of the proposed method across different model sizes.

\bibliographystyle{plain}




\end{document}